\documentclass[10pt,twocolumn,letterpaper]{article}

\usepackage{wacv}
\usepackage{times}
\usepackage{color}
\usepackage{epsfig}
\usepackage{graphicx}
\usepackage{amsmath}
\usepackage{amssymb}
\usepackage{subfigure}

\usepackage[pagebackref=true,breaklinks=true,colorlinks,bookmarks=false]{hyperref}

\wacvfinalcopy 

\def\wacvPaperID{961} 

\ifwacvfinal\fi
\begin{document}

\title{CowStallNumbers: A Small Dataset for Stall Number Detection \\of Cow Teats Images}

\author{Youshan Zhang \\
Yeshiva University, NYC, NY\\
{\tt\small youshan.zhang@yu.edu}
}

\maketitle
\ifwacvfinal\thispagestyle{empty}\fi

\begin{abstract}
  In this paper, we present a small cow stall number dataset named CowStallNumbers, which is extracted from cow teat videos with the goal of advancing cow stall number detection. This dataset contains 1042 training images and 261 test images with the stall number ranging from 0 to 60. In addition, we fine-tuned a ResNet34 model and augmented the dataset with the random crop, center crop, and random rotation. The experimental result achieves a 92\% accuracy in stall number recognition and a 40.1\%  IoU score in stall number position prediction. 

\end{abstract}

\section{Introduction}

In recent years, the dairy industry has expanded rapidly as the demand for milk and milk products grow. One of the critical aspects of dairy farming is the management of the herd's health and productivity. To maximize milk production and maintain the cows' health, it is necessary to ensure that the cows have a comfortable environment, appropriate feed, and regular health checks. 
We previously studied how to classify different categories of teats~\cite{zhangdairy2022,zhang2022separable}. However, these teat images were taken by phone and manually cropped. To save time, we set up a camera toward the teat-end areas~\cite{zhang2022unsupervised}; we proposed to extract the key teat frames from the video images, which leads to another problem that we cannot recognize the cows' ID unless we can recognize the stall number in the videos. Therefore, we need to develop a model to recognize the stall numbers.

\begin{figure*}[h]
\centering
\includegraphics[width=2 \columnwidth]{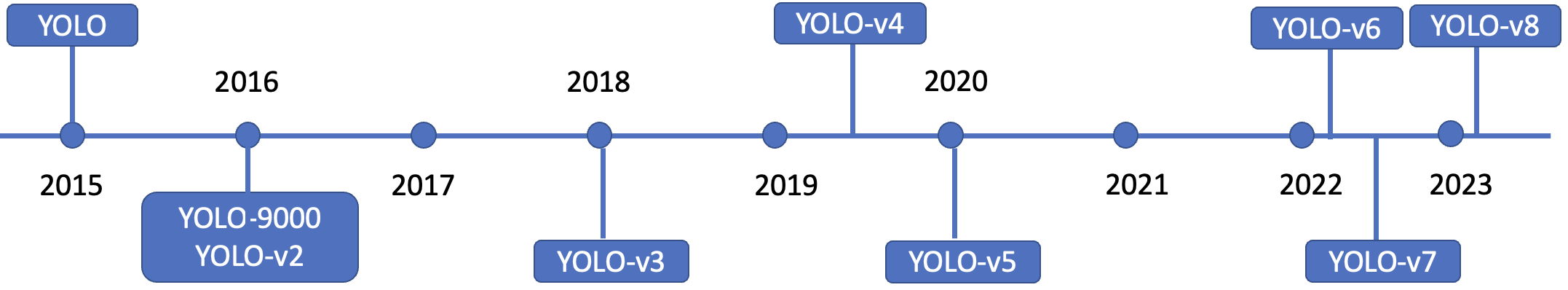}
\caption{Timeline of You Only Look Once (YOLO) variants.}
\label{fig:yolo}
\end{figure*}

In this paper, we fine-tune the ResNet34 model for cow stall detection. We evaluate our method using a small dataset of cow images with ground truth annotations for cow stall locations. Our results show that our method achieves high accuracy in cow stall detection,
and can provide a valuable tool for dairy farmers to optimize their operations. 
By enabling accurate and efficient cow stall detection, we can improve the health and productivity of cows and, in turn, increase the sustainability and profitability of the dairy industry. Our contributions are two-fold:
\begin{itemize}
    \item We present a small cow stall number
dataset named CowStallNumber with the goal of advancing the field of cow stall number detection. 
    \item We fine-tuned the pre-trained ResNet34 model to get the position of the stall number and recognize different stall numbers. We achieved a 95\% accuracy in stall number recognition and a 40.1\% IoU score in box position prediction.
\end{itemize}

\section{Related Work}\label{sec:related}
The faster Region-based Convolutional Neural Network (R-CNN) algorithm has been applied for cow tail detection, achieving higher accuracy and faster detection times than the R-CNN algorithm~\cite{huang2019cow}. However, one limitation of these object detection algorithms is that they struggle with detecting objects in certain environmental conditions, such as poor lighting or low contrast. This is a concern in cow detection since cows blend in with their surroundings or have their identifying features obscured by shadows or other visual obstructions.

To address this issue, thermal imaging cow detection has been explored~\cite{anagnostopoulos2021study}. Thermal imaging uses infrared technology to detect the heat signature of an object, which allows for more accurate detection in low-light or low-contrast environments. In addition, thermal imaging can detect cows even in situations where they are partially hidden, such as behind bushes or in tall grass. Despite the advances in cow detection technology, there are still challenges to be overcome. For example, detecting cows in large herds can be difficult, as cows can tightly pack together and partially obstruct each other from view. Furthermore, different breeds of cows may have unique physical characteristics that require specific detection algorithms.  We will review the related work of object detection in the following paragraphs. 


\begin{figure*}[h]
\centering  
\includegraphics[width=2 \columnwidth]{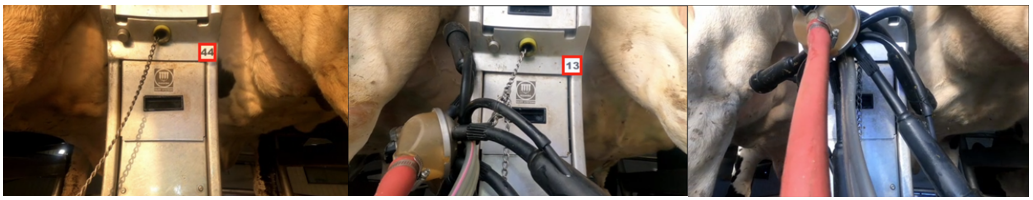}
\caption{Example of three stall images (44, 13, and 0).}
\label{fig:stall}
\end{figure*}

 \begin{figure*}[t!]
\centering
\includegraphics[width=2 \columnwidth]{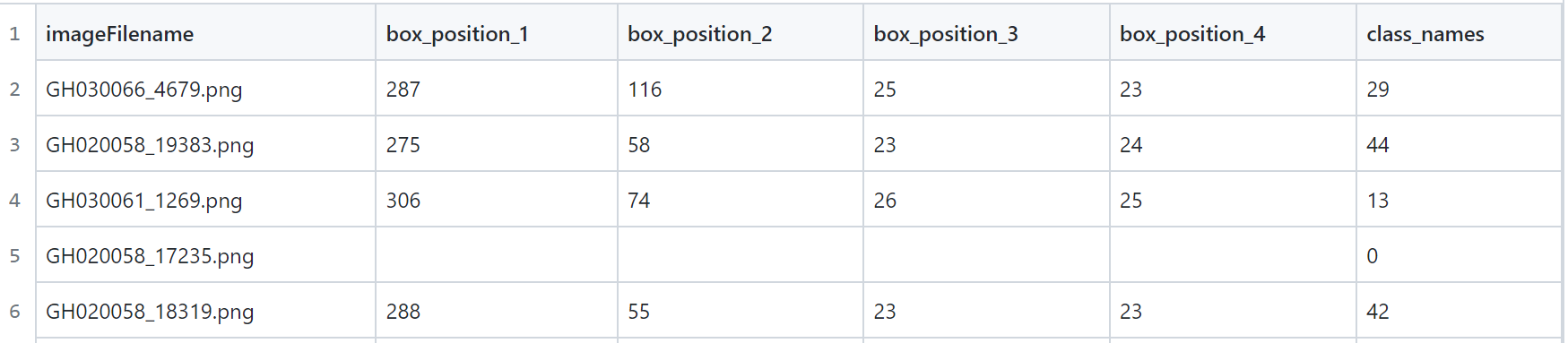}
\caption{Example contents in the CSV file. ImageFilename is the cow stall number image. [Box\_position\_1, Box\_position\_2, Box\_position\_3, Box\_position\_4] means that [x, y, height, width] of the bounding box location. Class\_name is the stall number class. Empty space means we cannot detect the stall number in the image.} 
\label{fig:csv}
\end{figure*}

  Girshick et al. proposed the R-CNN model, which applied a region proposal algorithm to generate candidate object regions. Then, it is passed to a convolutional neural network (CNN) for classification~\cite{girshick2015region}. Later, Girshick proposed a Fast R-CNN model, which is an improved version of R-CNN that uses a shared feature map for region proposal and classification, resulting in faster and more accurate detection~\cite{girshick2015fast}. Ren et al. developed a Faster R-CNN model, which is another extension of R-CNN that introduces a Region Proposal Network (RPN) to generate region proposals more efficiently, making it the first real-time object detection method~\cite{ren2015faster}. You Only Look Once (YOLO) is a real-time object detection method that generates predictions directly from the whole image, which is based on a single CNN that simultaneously predicts the class and position of each object. Redmon et al.~\cite{liu2016ssd} presented a Single Shot MultiBox Detector (SSD) model, which generates object proposals and predicts their class and location using a single neural network. SSD is faster compared to Faster R-CNN and achieves higher accuracy than YOLO. Lin et al. proposed a RetinaNet~\cite{lin2017focal}, which introduces a novel focal loss function that down-weights easy examples and up-weights hard examples. Mask R-CNN~\cite{he2017mask} is an extension of Faster R-CNN, which adds a branch to predict segmentation masks for each object, and it can segment the detected objects in addition to their bounding boxes. Cascade R-CNN~\cite{cai2018cascade} model is a variant of Faster R-CNN that utilizes a cascade of detectors to improve the detection quality. Lin et al. proposed a Feature Pyramid Network (FPN) model~\cite{lin2017feature} that generates a pyramid of feature maps that are used for object detection at different scales and improves the performance of object detectors. FPN. DetNet~\cite{li2018detnet} incorporates deformation-aware convolutional layers that can handle object deformations and viewpoint changes. DenseBox~\cite{huang2015densebox} applies dense regression to predict the coordinates of object bounding boxes.
CoupleNet~\cite{zhu2017couplenet} can predict the position and size of objects in a coupled manner, leading to improved performance compared to other real-time detection methods. Kong et al. proposed a Reverse Connection with Objectness Prior Networks (RON) model~\cite{kong2017ron} that uses a reverse connection to capture context information for object proposal generation. RefineDet~\cite{zhang2018single} is a multi-stage object detection method that progressively refines the location and size of object proposals. Edge Sampling Decoder (ESD)~\cite{chen2019towards} incorporates edge sampling into deep neural networks to detect edges and boundaries of objects. Neural Architecture Search with Feature Pyramid Network (NAS-FPN)~\cite{ghiasi2019fpn} uses neural architecture search to design a feature pyramid network that improves object detection performance. 

There are also different variants of YOLO methods as shown in Fig.~\ref{fig:yolo}. YOLOv1~\cite{redmon2016you} was introduced in 2016.  YOLOv2 (YOLO9000)~\cite{redmon2017yolo9000} was released in 2017 and used a Darknet-19 network architecture and incorporated batch normalization and anchor boxes. It also introduced multi-scale predictions, enabling the detection of objects at different scales.
YOLOv3~\cite{redmon2018yolov3} was released in 2018 and used a Darknet-53 network architecture, introduced feature pyramid networks (FPNs), and used residual connections. It also used anchor boxes with dynamic scaling, leading to better detection at different scales. YOLOv4~\cite{bochkovskiy2020yolov4}  was released in 2020, which used a CSPDarknet-53 network architecture, added various improvements such as spatial pyramid pooling, and used a complex data augmentation pipeline. YOLOv4 also included efficient training approaches such as mosaic data augmentation and focal loss.
YOLOv5~\cite{Joseph2020YOLOv5} was a PyTorch implementation rather than a fork from the original Darknet, which has a CSP backbone and PA-NET neck. The major improvement includes auto-learning bounding box anchors.
YOLOv6~\cite{li2022yolov6} was released in 2022 and was a single-stage object detection framework dedicated to industrial applications, with hardware-friendly efficient design and high performance. YOLOv7~\cite{wang2022yolov7} was released in 2022 and reduced the number of parameters by 75\%, requires 36\% less computation, and achieves 1.5\% higher AP (average precision) compared with YOLOv4. YOLOv7-X achieves a 21 FPS faster inference speed than YOLOv5-X. The YOLOv8~\cite{Joseph2023YOLOv8} model is fast, accurate, and easy to use, which can be applied to a wide range of object detection and image segmentation tasks. It can be trained on large datasets and run on various hardware platforms, from CPUs to GPUs.

\section{Data collection}\label{sec:data}
The stall number images are retrieved from cow teat videos, which are recorded to inspect the cow teats' health status. More details of video recording settings can be found in~\cite{zhang2022unsupervised}. We first applied the unsupervised few-shot key frame extraction (UFSKEF) model in~\cite{zhang2022unsupervised} to extract the coarse stall number key frames. We then manually checked these key frames and removed the wrong key frame images. Fig.~\ref{fig:stall} shows three example stall numbers, where 0 means that we cannot detect the stall numbers. Tab.~\ref{tab:stas} lists the statistics of our CowStallNumber dataset. Fig.~\ref{fig:csv} represents the content of training and test CSV files.

\begin{figure*}[t!]
\centering
\includegraphics[width=2 \columnwidth]{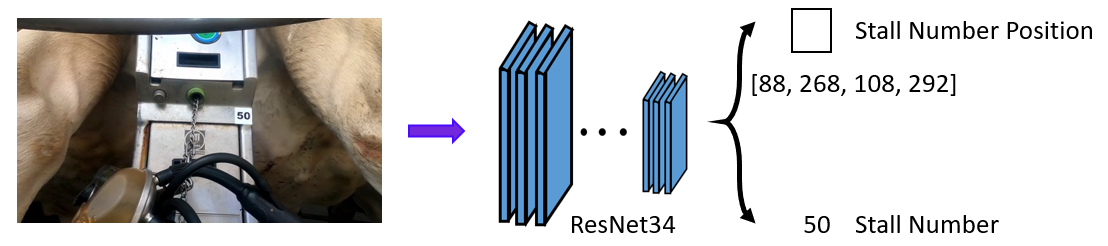}
\caption{The architecture of our stall number detection model. We fine-tune a ResNet34 model and add two output layers (for stall number box $[88, 268, 108, 292]$ and stall number $50$, respectively).}
\label{fig:model}
\end{figure*}

\begin{figure*}[h!]
\centering
\includegraphics[width=2 \columnwidth]{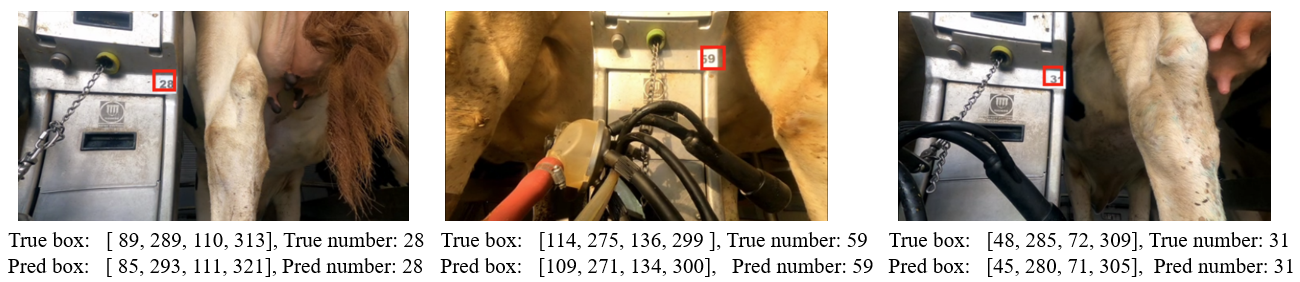}
\caption{Three example outputs of our model. True box: true bounding box location; Pred box: predicted bounding box location.}
\label{fig:output}
\end{figure*}

\begin{table}[h]
\small
\begin{center} 
\caption{Statistics of our CowStallNumber dataset}
 \setlength{\tabcolsep}{+3.3mm}{
\begin{tabular}{ccccc|c|c|c|c|c|c|c|}
\hline \label{tab:stas}
& \bf Training &  \bf Test & \bf Total \\
\hline
Numbers & 1042 & 261 & 1303\\
\hline
\end{tabular}}
\end{center}
\end{table}

\section{Methods}
Since stall number detection is an object detection problem, we evaluate the performance of the stall number box using the IoU score in Eq.\eqref{eq:IoU} and report the recognition accuracy of stall number using accuracy in Eq.\eqref{eq:acc}. 

\begin{equation}\label{eq:IoU}
     IoU=\frac{|A \cap B|}{|A \cup B|},
\end{equation}
where $A$ is the ground truth stall bounding box, and $B$ is the predicted stall bounding box.

\begin{equation}\label{eq:acc}
    \text{Accuracy} =\frac{1}{n} \sum_{i=1}^n (Y_i == \hat{Y_i}) \times 100,
\end{equation}
where $Y_i$ is ground truth stall number, and $\hat{Y_i} =f(X_i)$ is the predicted stall number ($X_i$ is the input stall image and $f$ is our fine-tune ResNet34 model).

We apply L1 loss to optimize the bounding box and cross-entropy loss to optimize the stall number recognition. The overall  architecture of our model is shown in Fig.~\ref{fig:model}, and the objective function is shown in the following equation. 
\begin{equation}
    \mathcal{L} =     \frac{1}{n} \sum_{i=1}^{n}  \{\alpha |A_i-B_i|_1 + \mathcal{L}_{ce} (f(X_i), Y_i) \}, 
\end{equation}
where $|\cdot|$ is the L1 loss, $\mathcal{L}_{ce}$ is the typical cross-entropy loss and $\alpha$ is the balance factor between L1 loss and cross-entropy loss.

\section{Implementation details}
During the training of our model, we set batch size = 64, training iteration $I = 300$, and learning rate = 0.06, $\alpha = 1/1000$ with an Adam optimizer on a 48G RTX A6000 GPU using PyTorch. The input image size is $270 \times 480$. We also perform the random crop, center crop, and random rotation to augment more images. The dataset is available at  \url{https://github.com/YoushanZhang/Cow_stall_number}.

\section{Results}\label{sec:results}
Tab.~\ref{tab:R2} shows the results of our model. We can find that the stall number recognition accuracy is high, while the IoU score is low. This implies that our model can successfully recognize the stall numbers. However, the low IoU score means that the bounding box of the stall number is not correctly predicted, which can lead to a low $\alpha$ value during the training. Another reason is that there are significant differences between different stall images (e.g., light, location, etc.). Fig.~\ref{fig:output} shows three output images of our model. All stall numbers are correctly predicted, while the stall number positions differ slightly from the ground truth positions.

\begin{table}[h]
\begin{center} 
\caption{Stall number detection results (\% means $\times 100$)}
 \setlength{\tabcolsep}{+3.3mm}{
\begin{tabular}{ccccc|c|c|c|c|c|c|c|}
\hline \label{tab:R2}
\bf Method &  \bf IoU (\%) & \bf Accuracy (\%)\\
\hline
ResNet34 & 40.1 & 95.0 \\
\hline
\end{tabular}}
\end{center}
\end{table} 

\vspace{-0.6cm}
\section{Conclusion}\label{sec:conclusion}
In this paper, we create a small cow stall number dataset named CowStallNumbers with the goal of advancing cow stall number detection. We fine-tuned a ResNet34 model and augmented the datasets with the  random crop, center crop, and random rotation. The experimental result achieves a
92\% accuracy in stall number recognition and a 40.1\% IoU
score in stall number position prediction. For future work, exploring a better object detection model (e.g., YOLOV8) could further improve the performance of stall number detection.

{\small
\bibliographystyle{plain}
\bibliography{egbib}
}

\end{document}